\documentclass[letterpaper]{article} 
\usepackage{appendix}
\usepackage{aaai24}  
\usepackage{times}  
\usepackage{helvet}  
\usepackage{courier}  
\usepackage[hyphens]{url}  
\usepackage{graphicx} 
\usepackage{natbib}  
\usepackage{caption} 
\usepackage{algorithm}
\usepackage{algorithmic}
\usepackage{booktabs}    
\usepackage{multirow}   
\usepackage{makecell}
\usepackage{array}
\usepackage[table]{xcolor}
\usepackage{amsfonts}
\usepackage{amsmath}
\usepackage{cite}
\usepackage{newfloat}
\usepackage{listings}
\DeclareCaptionStyle{ruled}{labelfont=normalfont,labelsep=colon,strut=off} 
\floatstyle{ruled}
\newfloat{listing}{tb}{lst}{}
\floatname{listing}{Listing}
\usepackage{tikz}
\usepackage{pgfplots}

\title{Zero-Shot Aerial Object Detection with Visual Description Regularization}

\author {
    Zhengqing Zang\textsuperscript{\rm 1,\rm 2}\equalcontrib,
    Chenyu Lin\textsuperscript{\rm 1,\rm 2}\equalcontrib,
    Chenwei Tang\textsuperscript{\rm 1,\rm 2},
    Tao Wang\textsuperscript{\rm 1,\rm 2}\thanks{Corresponding author},
    Jiancheng Lv\textsuperscript{\rm 1,\rm 2}
}
\affiliations {
    \textsuperscript{\rm 1}College of Computer Science, Sichuan University, Chengdu, 610065, P. R. China\\
    \textsuperscript{\rm 2}Engineering Research Center of Machine Learning and Industry Intelligence,\\
    Ministry of Education, Chengdu, 610065, P. R. China\\
    \{2022223045158, 2022223040017\}@stu.scu.edu.cn, tangchenwei@scu.edu.cn, twangnh@gmail.com, 
    lvjiancheng@scu.edu.cn
}

\begin{document}

\maketitle

\begin{abstract}

Existing object detection models are mainly trained on large-scale labeled datasets. However, annotating data for novel aerial object classes is expensive since it is time-consuming and may require expert knowledge. Thus, it is desirable to study label-efficient object detection methods on aerial images. In this work, we propose a zero-shot method for aerial object detection named visual Description Regularization, or \emph{DescReg}. 
Concretely, we identify the weak semantic-visual correlation of the aerial objects and aim to address the challenge with prior descriptions of their visual appearance. 
Instead of directly encoding the descriptions into class embedding space which suffers from the representation gap problem, we propose to infuse the prior inter-class visual similarity conveyed in the descriptions into the embedding learning. The infusion process is accomplished with a newly designed similarity-aware triplet loss which incorporates structured regularization on the representation space. 
We conduct extensive experiments with three challenging aerial object detection datasets, including DIOR, xView, and DOTA. The results demonstrate that DescReg significantly outperforms the state-of-the-art ZSD methods with complex projection designs and generative frameworks, e.g., DescReg outperforms 
best reported ZSD method on DIOR by 4.5 mAP on unseen classes and 8.1 in HM. 
We further show the generalizability of DescReg by integrating it into generative ZSD methods as well as varying the detection architecture.
Codes will be released at \url{https://github.com/zq-zang/DescReg}.
\end{abstract}


\section{Introduction}

\begin{figure}[t]
    \centering
    \includegraphics[width=\linewidth]{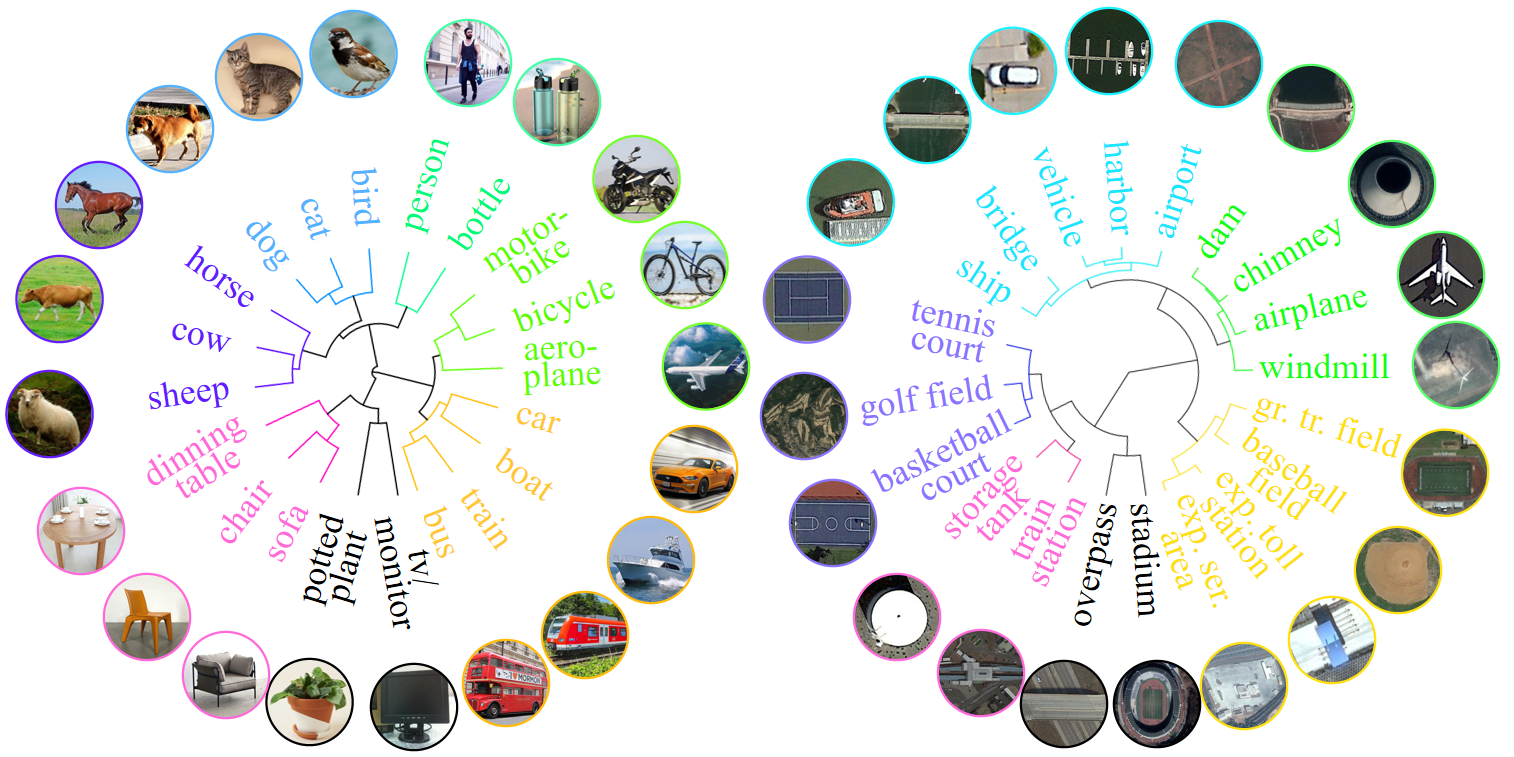}
    \caption{Illustration of weak semantic-visual correlation problem. We perform hierarchical clustering with semantic embeddings and show the radial dendrogram for the 20 common object classes from the Pascal VOC dataset (left) and the 20 aerial object classes from the DIOR dataset(right).
    The common object classes show clear clustering result which corresponds well to visual appearance (e.g., horse, cow, and sheep), while the semantic clustering of aerial object classes are inevident and shows much less correlation with visual appearance. Best viewed with zoom-in.}
    \label{fig:fig1:preliminary_analysis}
\end{figure}


Aerial object detection aims to detect objects from aerial images~\cite{xia2018dota,yang2019clustered,ding2021object}, e.g., images captured from an unmanned aerial vehicle (UAV). It plays an important role in many remote sensing applications, such as UAV-aided environmental monitor and disaster response systems.
Benefiting from the development of deep convolution neural networks (CNNs), aerial object detection has been extensively studied and advanced~\cite{yang2019clustered,zhu2021detection,li2022oriented,li2020density,deng2020global,han2021align} in recent years. Prior research mainly focuses on improving the accuracy or efficiency based on a fully supervised paradigm. However, labeling objects for large-scale aerial images is extremely costly due to the small object size and irregular viewing angle.
Hence expanding the vocabulary becomes a challenge for fully supervised aerial object detection methods~\cite{Lam2018xViewOI}.

\emph{Zero-shot object detection} (ZSD), which aims to detect unseen object classes without bounding box annotations~\cite{Bansal2018ZeroShotOD,Demirel2018ZeroShotOD,rahman2018zero}, appears as a promising approach for reducing the copious label demand in aerial object detection. ZSD methods mainly leverage the semantic relation between object classes to detect unseen classes, e.g., Cat and Dog are semantically similar, and thus the knowledge learned on Cat could be transferred to recognize Dog. Methodologically, this knowledge transfer process is typically realized through learning a command embedding function to align visual and semantic features~\cite{rahman2018zero,Bansal2018ZeroShotOD,Zheng2020BackgroundLC,rahman2020improved}, or learning a universal synthesizer function to generate training samples~\cite{Hayat2020SynthesizingTU,Zhao2020GTNetGT,Zhu2019DontEL,Huang2022RobustRF,sarma2022resolving}.
However, we find that existing ZSD methods perform poorly on aerial images due to weak semantic-visual correlation. Concretely, as shown in Fig.~\ref{fig:fig1:preliminary_analysis}, our core observation is that objects in natural images tend to be visually distinct and align well with semantic clustering, yet objects from aerial images often appear vague and lack semantic correlation. Such an issue hinders effective recognition of unseen classes.


Based on the analysis, we aim to introduce additional textual descriptions to augment the semantic space of aerial object classes. These descriptions serve as prior knowledge and provide details on the visual characteristics. One straightforward way to utilize these descriptions is to encode them into semantic embeddings, e.g., through a pre-trained language model. We find such encoding improves the performance but the gain is limited, likely due to the representation gap problem of visual-semantic space~\cite{wang2017zero}, which is more severe in aerial images. 
Based on this finding, we instead leverage the textual description information as a structural regularization and propose a description regularization method, or \emph{DescReg}.
The intuition is to preserve the visual similarity structure in the classification space and thus better transfer the knowledge from seen classes to unseen classes. 
To instantiate the structural regularization, we develop an adaptive triplet loss where each projected class embedding is treated as independent samples. Specifically, The positive pairs are sampled from similar classes, and negative pairs from dissimilar classes. Then the difference between the positive pair and the negative pair is employed as the margin. Such a triplet loss formulation is similarity-aware and helps effectively preserve the inter-class similarity relation in the embedding during optimization. 



To validate the above method, we establish two challenging zero-shot aerial object detection setups with DOTA and xView datasets. Together with the existing aerial ZSD setup on the DIOR dataset~\cite{Huang2022RobustRF}, we conduct extensive experiments on the two-stage Faster R-CNN detector and further show generalization the multi-stage Cascaded R-CNN detector~\cite{cai2018cascade} and the popular one-stage YOLOv8 detector~\cite{redmon2017yolo9000,Jocher_YOLO_by_Ultralytics_2023}. DescReg effectively improves the detection accuracy of raw baseline method on both seen and unseen classes. Remarkbaly, DescReg with simple one-layer projection outperforms the SOTA generative ZSD methods~\cite{Huang2022RobustRF} by 4.5 in unseen mAP and 8.1 in HM, with the same detection architecture. We further incorporate our method into the generative ZSD method by regularizing the visual feature synthesizing process and observe significant improvement, which demonstrates the strong generalizability of our DescReg as a structural similarity regularization method.

In summary, Our contributions are four-fold:
\begin{itemize}
    \item To the best of our knowledge, we are the first to conduct a comprehensive study on zero-shot aerial object detection, with in-depth analysis and focused methodological design.
    \item We identify the weak semantic-visual correlation challenge in aerial image domain and propose to leverage prior visual text descriptions to address the challenge. 
    \item To effectively incorporate the visual cues conveyed in the textual descriptions, we propose a novel inter-class similarity-aware triplet loss to encode the inter-class relation as a structural regularization.
    \item We established two new challenging ZSD setups with the DOTA and xView datasets and conduct extensive experiments with various detection architectures to validate and understand the proposed method.
    
\end{itemize}

\section{Related Work}
\subsection{Zero-shot Object Detection}

Motivated by studies on zero-shot learning(ZSL), which aims to recognise unseen classes by transferring semantic knowledge from the seen ones\cite{Mishra2018AGA,tang2019angle,Jasani2019SkeletonBZ, Demirel2019LearningVC, tang2020zero,tang2021zero}, the more challenge task of zero-shot detection(ZSD) proposed in 2018 \cite{Bansal2018ZeroShotOD} has received widespread attention recently. The purpose of ZSD is not only to categorize the unseen objects but also localize them. Quite similar to ZSL task, approaches to solving this problem focus on either embedding-based approaches or generative-based approaches. Embedding-based ZSD aims to learn a visual$\rightarrow$semantic projection from seen classes to promote the alignment of two spaces\cite{Demirel2018ZeroShotOD, Li2019ZeroShotOD}. In order to better distinguish between background and unseen classes, background vector was refined by learnable parameters in \cite{Zheng2020BackgroundLC}. 

Generative-based methods tackle this problem from the other paradigm\cite{Zhu2019DontEL, Zhao2020GTNetGT}. They train a generative adversarial network(GAN) to synthesize visual samples of unseen classes, thereby obtaining unseen classifiers and regressors under supervised paradigms. To maintain inter-class structure and increase intra-class diversity, \cite{Huang2022RobustRF} proposed a robust region feature synthesizer with two novel components. For the same purpose. \cite{sarma2022resolving} proposed a triplet loss computed by visual features and added a cyclic consistency loss to constrain visual-semantic alignment. However, none of these methods have recognized the important role of category embedding representation in the performance of ZSD, which is the focus of this study.



\subsection{Aerial Object Detection}


Aerial images refer to images captured by sensors mounted on satellites, aircraft, or drones, which are used to observe and collect information about the Earth’s surface from a distance.
Despite the huge progress made in object detection tasks on natural images, The aerial object detection task remains challenging.
Prior studies on aerial object detection mainly focus on handling the innate challenge of aerial imagery such as small target size variation~\cite{recasens2018learning,yang2019clustered,meethal2023cascaded,li2020density,yang2022querydet,koyun2022focus} and object rotation~\cite{Cheng2016LearningRC, Zhang2020RotationInvariantFL, Cheng2019LearningRA}.
Despite these efforts, few works study label-efficient detection in aerial images. Some studies~\cite{wolf2021double,lu2023few} propose few-shot learning approaches for aerial object detection, however, these methods still require target object labels.
In this work, we study how to leverage existing training data on seen classes and directly generalize to arbitrary unseen classes, i.e., zero-shot detection.





\section{Proposed Method}

\begin{figure*}[htbp]   
	\centering
	\includegraphics[width=\linewidth,scale=1.00]{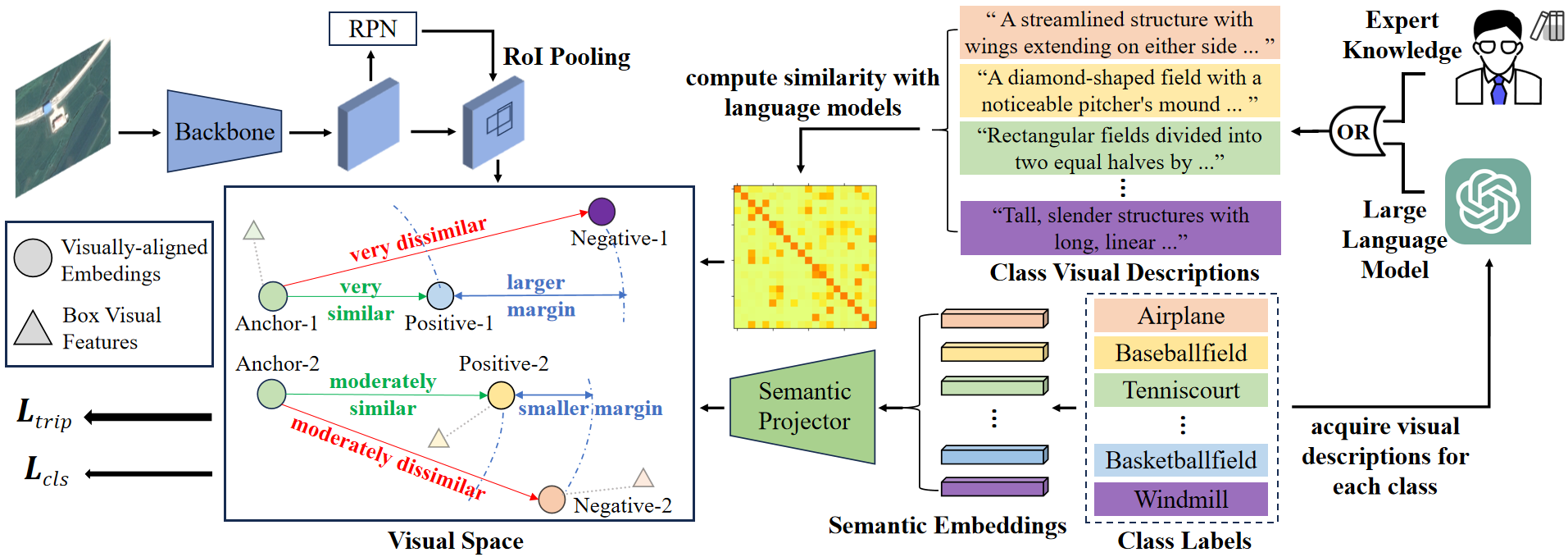}
	\caption{The overall framework of the proposed method.}
	\label{Figure:figure_overall}
\end{figure*}

\subsection{Overview}

Given the bounding box annotation on a set of seen object categories $\mathbb{F} = \{C_1, C_2,... C_N\}$, zero-shot object detection (ZSD) aims at training on the seen data and generalizing to a set of target unseen object categories $\mathbb{F}^*=\{C_1, C_2,... C_M\}$. In the following paragraphs, we first present our main detection architecture and then introduce our approach with in-depth analyses under the context of ZSD.

\paragraph{Object Detection Architecture}
The classical two-stage Faster R-CNN~\cite{ren2015faster} detection model consists of a visual feature extraction backbone $\mathcal{T}$, a region proposal network $\mathcal{R}$, a shared multi-layer feature transformation network $\mathcal{F}$. a region classifier $\mathcal{C}$, and a box regressor $\mathcal{B}$. Given input image $\textbf{I}$, the model first extracts image feature $\textbf{F}$: $\textbf{F}=\mathcal{T}(\textbf{I})$. Then object candidate proposals are predicted by the region proposal network: $\{\textbf{p}_i\}=\mathcal{R}(\textbf{F})$. With the proposals, feature pooling is conducted on the image feature map $\textbf{F}$ to obtain the proposal region feature $\{\textbf{v}_i\}$. The feature is then further refined by the shared network: $\textbf{v}_i=\mathcal{F}(\textbf{f}_i)$. Finally, object classification scores and refined bounding boxes are predicted by the classifier and the regressor: $\textbf{s}_i=\mathcal{C}(\textbf{v}_i)$, $\textbf{b}_i=\mathcal{B}(p_i, \textbf{v}_i)$.

The detection model can be trained on the seen class data with proposal loss, classification loss, and regression loss: $\mathcal{L}=\mathcal{L}_{prop}+\mathcal{L}_{cls}+\mathcal{L}_{reg}$. The trained region proposal network is class-agnostic and thus may generalize directly to predict the unseen classes. The box regressor is also not sensitive to classes and thus can be applied directly to unseen classes, i.e., by using the class-agnostic version or using prediction from seen classes~\cite{Huang2022RobustRF}. The major challenge here is to generalize the classification to unseen classes, as the region classifier is only trained on the seen class data and cannot predict the unseen classes. 

\paragraph{Detecting the Unseen with Semantic Bridging}
While unseen class data is not available, the semantic relation can be efficiently represented with semantic word embeddings. These embeddings can be obtained from pre-trained word embedding models such as Word2vec~\cite{mikolov2013distributed} and large language models such as BERT~\cite{devlin2018bert}: $\textbf{c}_j=\mathcal{W}(C_j)$, where $\textbf{c}_j$ is the vectorized representation and $\mathcal{W}$ is the embedding model. With these embeddings and trained detection models on the seen class data, existing zero-shot object detection methods mainly focus on bridging the gap between seen and unseen classes. These methods can be classified into \emph{embedding-alignment} and \emph{generative} methods. The embedding-alignment methods~\cite{khandelwal2023frustratingly,zhang2020zero,yan2022semantics}
aim to bridge the gap between visual and semantic space by learning representation alignment. 
For example, learning an alignment function $\phi$ to align semantic embeddings to visual features~\cite{zhang2020zero}:
\begin{align}
    \textbf{w}_j=\phi(\textbf{c}_j)
\end{align}
where $w_j$ is the visually-aligned class representation. With $w_j$, the visual features $v_i$ can be classified based on similarity metrics such as cosine similarity, and the seen classification supervision is employed to learn the alignment function. The learned alignment function is expected to generalize to unseen classes by utilizing unseen class embeddings, and thus the detection model can detect unseen objects.

\paragraph{The Semantic-Visual Correlation Challenge}
Although such embedding-alignment methods are shown to be effective on natural image datasets such as Pascal~\cite{everingham2010pascal} and COCO~\cite{lin2014microsoft}. 
They suffer from poor semantic-visual correlation on aerial images: \emph{the semantic embedding $\textbf{c}_j$ has poor correlation with visual features $\textbf{v}_i$, which leads to severe difficulty for the learned embedding function $\phi$ to generalize on the unseen classes.} Note this observation also applies to the generative methods which will be discussed in the next section.

Based on this observation, we aim to improve the semantic-visual correlation by augmenting the semantic embeddings with extra visual cues. The visual cues are instantiated as simple textual descriptions. This is motivated by the prior works in zero-shot learning that employs textual descriptions to augment the recognition of unseen classes~\cite{elhoseiny2013write,paz2020zest}. The descriptions can be obtained from \emph{experts} or simply through large-scale pre-trained \emph{large language models} (LLM), e.g., GPT~\cite{OpenAI2023}.

\subsection{DescReg Formulation}

The textual descriptions are free-form and efficient to acquire, they can help provide valuable information such as shape, color, and context for the aerial objects. 
However, we find simply encoding them into the semantic representation offers limited gain. This is likely due to the following issues:  \textbf{1)} the representations of semantic feature space and visual feature space have distinct distributions, which causes ineffective transformation of the descriptions into visual feature space.
\textbf{2)} The image feature representations of aerial objects are less discriminative due to their smaller size than that of common objects, thus it is more difficult to classify them against the false classes and backgrounds.

To address the above issues, we leverage the inter-class visual similarity information as a structural regularization to learn more discriminative alignment functions. Specifically, given the visual descriptions for all seen and unseen classes: $\{T_j\}$, pre-trained language models such as BERT~\cite{devlin2018bert} are used to encode them into vectorized representation:
\begin{align}
    \textbf{t}_j=\mathcal{W}(T_j)
\end{align}
where $\mathcal{W}$ is the employed language model and $\textbf{t}_j$ is the obtained representation.
Then we compute the pair-wise cosine similarity and obtain the similarity matrix \textbf{S}:
\begin{align}
    \textbf{S}(j, k)=\frac{\textbf{t}_j\textbf{t}_k}{||\textbf{t}_j||_2||\textbf{t}_k||_2}
\end{align}
To encourage more discriminative inter-class similarity and keep the similarity score within the value range $(0,1]$, we introduce a self-excluding Softmax:
\begin{align}
\hat{\textbf{S}}(j, k) =
    \begin{cases}
    \frac{e^{\textbf{S}(j, k)/\tau }}{\sum_{k^{'}\neq j}e^{{\textbf{S}(j, k^{'})/\tau}}} & \text{if}\ k\neq j\\
    \textbf{S}(j, k) & \text{otherwise}
    \end{cases}
\end{align}
where $\hat{\textbf{S}}$ is the normalized similarity matrix, of which all the diagonal elements are $1$, corresponding to self-similarity and the other elements are in the value range $(0,1)$, corresponding to inter-class similarity.

The visual characteristics of object classes are now encoded structurally as this similarity matrix. We then integrate it into the embedding alignment learning process. Motivated by the triplet loss~\cite{frome2013devise,akata2015label}, we treat the visually-aligned semantic representations $\textbf{w}_j$ as independent feature samples and perform positive-negative sampling based on the similarity, then triplet loss is imposed on the samples:
\begin{align}
    \mathcal{L}^j_{trip}=max\{0, d(\textbf{w}_j, \textbf{w}_{h(j)})-d(\textbf{w}_j, \textbf{w}_{l(j)})+\Delta\}
\end{align}
where $d(\cdot,\cdot)$ is the Euclidean distance. $h(j)$ denotes sampling a similar class for class $j$ and $l(j)$ means sampling a less similar class, or dissimilar class. The sampling is conducted based on similarity scores $\hat{\textbf{S}}(j,:)$. $\Delta$ is the margin. However, such a direct adoption of triplet loss does not consider the similarity level between classes, e.g., a bridge may be very similar to a dam, but less similar to an overpass, while a vehicle may look a bit dissimilar to a boat but is very distinct to a baseball-field. We thus propose to employ the similarity gap as the margin for the triplet regularization:
\begin{align}
    \Delta_j=\hat{\textbf{S}}(j,h(j))-\hat{\textbf{S}}(j,l(j))
\end{align}
such a second-order metric helps encode the discrepancy in similarity level into the margin regularization. It facilitates the structural learning of the alignment function and thus the knowledge learned from seen classes can better transfer to the unseen classes. The improved similarity-aware triplet loss is thus:
\begin{align}
    \mathcal{L}^j_{trip}=max\{0, d(\textbf{w}_j, \textbf{w}_{h(j)})-d(\textbf{w}_j, \textbf{w}_{l(j)})+\Delta_j\}
\end{align}
Unlike prior works that apply contrastive objectives~\cite{Huang2022RobustRF,yan2022semantics}, the proposed margin-adaptive triplet loss is less greedy and allows strong flexibility in representation space.
The loss is summed over all the seen and unseen classes to compute the full regularization objective:
\begin{align}
    \mathcal{L}_{trip}=\sum_{j}\mathcal{L}^j_{trip}
\end{align}

During the learning of the alignment function, the classification objective (e.g., cross-entropy) is usually computed on the seen classes. So the complete objective with DescReg is:
\begin{align}
    \mathcal{L}=\mathcal{L}_{cls}+\mathcal{L}_{trip}
\end{align}
Fig.~\ref{Figure:figure_overall} shows the overall framework integrated with Faster R-CNN.

\paragraph{Generalization to Generative Methods}
Unlike the above embedding-alignment methods, the generative methods~\cite{Hayat2020SynthesizingTU,Zhu2019DontEL,Huang2022RobustRF,rahman2020improved} aim to learn universal visual feature synthesizers. The method can be simplified as generating visual feature samples based on semantic embeddings:
\begin{align}
    \hat{\textbf{v}}_j=\varphi(\textbf{c}_j, \textbf{z})
\end{align}
where $\hat{\textbf{v}}_j$ is the synthesized visual feature and $\textbf{z}$ is random noise to encourage the feature diversity. The synthesized features can be employed to train the classifier for both seen and unseen classes. Similar to the above-mentioned semantic-visual correlation challenge, here the synthesizer also faces generalization issues on the unseen classes. The proposed similarity-aware triplet loss can then easily added to the training process of generative networks:
\begin{align}
    \mathcal{L}^j_{trip}=max\{0, d(\hat{\textbf{v}}_j, \hat{\textbf{v}}_{h(j)})-d(\hat{\textbf{v}}_j, \hat{\textbf{v}}_{l(j)}+\Delta_j\}
\end{align}

\begin{table*}[t]
\centering
\begin{tabular}{c|ccc|c|ccc|ccc}
   \hline
   \multirow{3}*{\textbf{Method}} & \multicolumn{4}{c}{\textbf{ZSD}} 
   \vline & \multicolumn{6}{c}{\textbf{GZSD}} \\
   \cline{2-11}
   & \multicolumn{3}{c}{\textbf{Recall@100}} 
   \vline & \multirow{2}*{\textbf{mAP}} &
   \multicolumn{3}{c}{\textbf{Recall@100}} 
   \vline & \multicolumn{3}{c}{\textbf{mAP}} \\
   \cline{2-4} \cline{6-11}
   & IoU=0.4 & IoU=0.5 & IoU=0.6 & &S & U & HM & S & U & HM \\
   \hline
   \rowcolor{lightgray!30} 
   BLC~\cite{Zheng2020BackgroundLC}  & - & - & - & - & - & - & - & 6.1 & 0.4 & 0.8 \\
   SU~\cite{Hayat2020SynthesizingTU}  & - & - & - & 10.5 & - & - & - & 30.9 & 2.9 & 5.3 \\
   \rowcolor{lightgray!30} 
   RRFS~\cite{Huang2022RobustRF} & - & - & - & 11.3 & - & - & - & 30.9 & 3.4 & 6.1 \\
  V2S$\dagger$~\cite{khandelwal2023frustratingly} & 14.1 & 11.9 & 10.1 & 4.1 & 78.2 & 15.8 & 26.3 & 57.0 & 1.4 & 2.7 \\
  \rowcolor{lightgray!30} 
  RRFS$\dagger$~\cite{Huang2022RobustRF} & 22.1 & 19.8 & 18.1 & 9.7 & 60.0 & 19.9 & 29.9 & 41.9 & 2.8 & 5.2 \\
  ContrastZSD$\dagger$~\cite{yan2022semantics} & 24.9 & 22.3 & 20.1 & 8.7 & 69.2 & 25.9 & 37.7 & 51.4 & 3.9 & 7.2 \\
  \rowcolor{lightgray!30} 
   DescReg (ours) & \textbf{37.9} & \textbf{34.6} & \textbf{31.5} & \textbf{15.2} & \textbf{82.0} & \textbf{34.3} & \textbf{48.4} & \textbf{68.7} & \textbf{7.9} & \textbf{14.2} \\
   \hline
\end{tabular}

\caption{Comparison with state-of-the-art methods under ZSD and GZSD settings on DIOR dataset. $\dagger$ denotes our implementation results. "S" and "U" denote seen classes and unseen classes, respectively.}
\label{tab:tab1_dior_results}
\end{table*}

\section{Experiments}
We study four questions in experiments. 1) How does DescReg improve the performance of zero-shot aerial object detection? is it efficient? 
2) Is DescReg sensitive to visual descriptions and embedding generation methods? 3) How does each component take effect? 
4) Can DescReg generalize to the generative ZSD methods and be applied on different object detection meta-architectures?

\subsection{Datasets and Experiment Setup}
We evaluate the proposed method on three challenging remote sensing image object detection datasets: DIOR~\cite{Li2019ObjectDI}, xView~\cite{Lam2018xViewOI}, and DOTA~\cite{Xia2017DOTAAL}.
For DIOR, we follow the setting in prior work~\cite{Huang2022RobustRF}. For xView and DOTA, we conduct semantic clustering and sample classes within clusters to ensure unseen class diversity and semantic relatness\cite{rahman2018zero,Huang2022RobustRF}. The resulting xView contains 48 seen classes and 12 unseen classes, and the resulting DOTA contains 11 seen classes and 4 unseen classes. We also perform cropping on the xView and DOTA images to simplify the data. Due to space limits, please refer to our supplementary file for more details. Throughout the experiments, unless otherwise stated, we adopt the Faster R-CNN model as the base detection model and IOU=0.5 for the evaluation.





\begin{table*}[htb]
\centering
\fontsize{9.5pt}{10.5pt}\selectfont
\setlength{\tabcolsep}{0.78mm}
\begin{tabular}{c|ccc|c|ccc|ccc|ccc|c|ccc|ccc}
   \hline
    & \multicolumn{10}{c}{\textbf{xView}}\vline & 
    \multicolumn{10}{c}{\textbf{DOTA}} \\
    \cline{2-21}
    \multirow{3}*{\textbf{Method}} & \multicolumn{4}{c}{\textbf{ZSD}}\vline & \multicolumn{6}{c}{\textbf{GZSD}}\vline & \multicolumn{4}{c}{\textbf{ZSD}}\vline & \multicolumn{6}{c}{\textbf{GZSD}} \\
   \hline
   & \multicolumn{3}{c}{\textbf{RE@100}}\vline & \multirow{2}*{\textbf{mAP}}
   & \multicolumn{3}{c}{\textbf{RE@100}}\vline & \multicolumn{3}{c}{\textbf{mAP}}
   \vline
   & \multicolumn{3}{c}{\textbf{RE@100}}\vline & \multirow{2}*{\textbf{mAP}} 
   & \multicolumn{3}{c}{\textbf{RE@100}}\vline & \multicolumn{3}{c}{\textbf{mAP}} \\
   \cline{2-4}\cline{6-8}
   \cline{9-11}\cline{12-14}\cline{16-18}
   \cline{19-21}
   & 0.4 & 0.5 & 
   0.6 & & S & U & HM & S & U & HM &
    0.4 & 0.5 & 
    0.6 & & S & U & HM & S & U & HM \\
    \hline
    \rowcolor{lightgray!30} 
    RRFS & 17.6 & 14.3 & 11.3 & 2.2 & 19.1 & 5.8 & 8.9 & 10.2 & 1.6 & 2.8 & 17.5 & 14.4 & 11.5 & 2.9 & 71.4 & 14.2 & 23.7 &
    47.1 & 2.2 & 4.2 \\
    ContrastZSD & 29.0 & 27.1 & 25.9 & 4.1 & 27.6 & 13.9 & 18.5 & 16.8 & 2.9 & 4.9 & 28.7 & 25.4 & 23.9 & 6.0 & 69.1 & 12.2 & 20.7 &
    41.6 & 2.8 & 5.2 \\
    \rowcolor{lightgray!30} 
    DescReg & \textbf{45.9} & \textbf{43.0} & \textbf{40.1} & \textbf{8.3} & \textbf{28.0} & \textbf{12.8} & \textbf{17.6} & \textbf{17.1} & \textbf{5.8} & \textbf{8.7} & \textbf{37.3} & \textbf{34.4} & \textbf{29.6} & \textbf{8.5} & \textbf{83.8} & \textbf{29.9} & \textbf{44.0} &
    \textbf{68.7} & \textbf{4.7} & \textbf{8.8} \\
   \hline
\end{tabular}

\caption{Performance of our proposed model on xView and DOTA datasets for ZSD and GZSD settings.}

\label{tab:tab2_xview_dota_results}
\end{table*}

\subsection{Implementation Details}
Following prior works~\cite{yan2022semantics,Huang2022RobustRF,yan2022semantics}, we adopt Faster R-CNN with ResNet-101~\cite{he2016deep} as the base detection architecture and conduct two-stage training. In the first stage, the model is first trained on the seen class data as conventional detection training, In the second stage, the model is frozen and the semantic-visual projection is fine-tuned with the proposed DescReg.
In addition to Faster R-CNN, we also validate our method on the newly released one-stage YOLOv8 model~\cite{Jocher_YOLO_by_Ultralytics_2023} and the cascaded detection model~\cite{cai2018cascade}. Due to space limit, please refer to supplementary for more details on implementation.


\subsection{Main Results}

\paragraph{Comparison with State-of-the-arts on DIOR}
In Tab.~\ref{tab:tab1_dior_results}, we compare the results with state-of-the-art methods on the DIOR dataset. The proposed method outperforms all compared methods in both ZSD and GZSD settings.
\emph{Under the ZSD setting}, our method achieves more than 11.0\% absolute gain for recalls of different IOU thresholds, and nearly 4.0\% mAP increase compared to the best-reported method, demonstrating its much stronger ability to detect unseen categories compared to All other ZSD methods. 
\emph{Under the GZSD setting}, the proposed method achieves the best mAP performance on seen classes, surpassing the best-compared method by 11.7\% in mAP, this result shows that our zero-shot learning method achieves the least interference on the seen class recognition. Furthermore, our method achieves 7.9\% unseen mAP and 14.2\% HM, which also significantly outperforms the prior methods.
Similar observations hold on the recall metrics.


\begin{table*}[htb]
\centering
\fontsize{9.5pt}{10.5pt}\selectfont
\setlength{\tabcolsep}{0.78mm}
\begin{tabular}
{c|c|c|c|c|c|c|c|c|c|c|c|c|c|c|c|c|c|c|c|c|c}
   \hline
    &  & \multicolumn{4}{c}{\textbf{DIOR}}\vline & 
    \multicolumn{4}{c}{\textbf{DOTA}}\vline & 
    \multicolumn{12}{c}{\textbf{xView}} \\
    \cline{3-22}
    \rotatebox{90}{\textbf{Setting}} & \textbf{Method} & \rotatebox{90}{airport}  
    & \rotatebox{90}{\makecell[c]{bask. f. }} & \rotatebox{90}{\makecell[c]{gr. tra. f.}}
    &  \rotatebox{90}{windmill} &\rotatebox{90}{\makecell[c]{tenn. c. }} & \rotatebox{90}{\makecell[c]{heli.}} & \rotatebox{90}{\makecell[c]{soccer.\\field}} & \rotatebox{90}{\makecell[c]{swim.\\pool}} & \rotatebox{90}{\makecell[c]{heli.}}
    & \rotatebox{90}{bus} & \rotatebox{90}{\makecell[c]{pic. track}}
    & \rotatebox{90}{\makecell[c]{tru. tra.\\w/ tox tra. }} & \rotatebox{90}{\makecell[c]{mar.\\vessel}} & \rotatebox{90}{motorb. } & \rotatebox{90}{barge} & \rotatebox{90}{\makecell[c]{reach\\stacker}}
    & \rotatebox{90}{\makecell[c]{mobile\\crane}} & \rotatebox{90}{scraper} 
    & \rotatebox{90}{excavator} & \rotatebox{90}{\makecell[c]{ship. cont.}}\\
    
    \hline
    \rowcolor{lightgray!30}
    \cellcolor{white} & RRFS & 3.1 & 2.0 & 6.3 & 0.0 & 4.4 & 0.0 & 4.5 
    & 0.0 & 0.8 & 2.0 & 0.0 & 0.0 & 3.9 & 5.5 & 5.2 & \textbf{0.1} &
    0.0 & \textbf{1.1} & 0.0 & \textbf{0.4}\\
    & ContrastZSD & \textbf{5.2} & 2.1 & 8.1 & 0.0 & 3.5 & \textbf{2.9} & 4.8 & 0.0 & 8.1 & \textbf{5.5} & \textbf{0.1} & 1.2 & 6.3 & \textbf{9.7} & 1.2 & 0.0 & 0.0 & 0.1 & 3.1 & 0.0\\ 
    \rowcolor{lightgray!30}

    \multirow{-3}{*}{\cellcolor{white}\rotatebox{90}{\textbf{GZSD}}}
     & DescReg & 0.0 & \textbf{9.2} & \textbf{20.0} & \textbf{2.4} & \textbf{9.1} & 0.1 &\textbf{9.5} 
    & 0.0 & \textbf{21.9} & 4.5 & 0.0 & \textbf{6.1} & \textbf{13.3} & 9.1 & \textbf{8.2} & 0.0 &
    0.0 & 0.4 & \textbf{6.1} & 0.0\\ 
    
    \hline
    & RRFS & \textbf{12.3} & 6.2 & 19.7 & 0.6 & 5.4 & 0.1 & 6.1 
    & 0.0 & 0.1 & 2.4 & 0.0 & 0.1 & 6.9 & 1.5 & 9.2 & \textbf{0.5} & 
    0.0 & \textbf{5.1} & 0.0 & 0.0\\
    \rowcolor{lightgray!30}
    \cellcolor{white} & ContrastZSD & 9.7 & 3.9 & 21.2 & 0.1 & 7.4 
    & \textbf{4.5} & 11.9 & 0.0 & 14.1 & \textbf{5.7} & 0.0 & 2.8 & 7.3 & \textbf{9.7} &
    1.1 & 0.1 & 0.0 & 0.6 & 7.6 & 0.0\\
    \multirow{-3}*{\rotatebox{90}{\textbf{ZSD}}} & DescReg & 0.1 & \textbf{10.9} & \textbf{45.7} & \textbf{3.9} & \textbf{11.3} & 0.4 & \textbf{22.2} & \textbf{0.1} & \textbf{36.0} & 4.9 & 0.1 & \textbf{7.5} & \textbf{19.8} & 9.1 & \textbf{10.2} & 0.0 &
    \textbf{0.4} & 0.9 & \textbf{10.3} & 0.0\\
   \hline
\end{tabular}

\caption{Class-wise AP comparison of different methods on unseen classes of three aerial image datasets.}

\label{tab:tab3_clswise_results}
\end{table*}

\paragraph{Experiments on xView and DOTA}
In addition to DIOR, we further conduct zero-shot detection experiments on the challenging xView and DOTA datasets. We compare to RRFS~\cite{Huang2022RobustRF} and ContrastZSD~\cite{yan2022semantics} as representatives of SOTA generative methods and embedding-alignment methods.
The result is shown in Tab.~\ref{tab:tab2_xview_dota_results}. On both datasets, our method shows higher performances compared to the baselines. 
Specifically,
Under both ZSD and GZSD settings of xView, the proposed method achieves nearly two-fold improvement in unseen mAP compared to the best-performing ContrastZSD method (4.1\% to 8.3\%, 2.9\% to 5.8\%). The corresponding gains on DOTA are about 50\% relatively.
We also observe that with xView, ContrastZSD achieves similar or higher recalls on the GZSD setting compared to our method, but the unseen mAP is lower, which indicates its unseen images may be less discriminative against the background, and thus predicts more false positives.

\paragraph{Class-wise Resutls}
We also report the class-wise mAP performance in terms of ZSD and GZSD for all three aerial object detection datasets. The results are shown in Tab.~\ref{tab:tab3_clswise_results}.
We note some unseen classes are very challenging and show near 0\% AP on the test set (e.g. 0.1\% and 0.4\% for helicopters on DOTA, under ZSD and GZSD settings respectively). This phenomenon is also observed in prior ZSD works~\cite{yan2022semantics,Huang2022RobustRF,yan2022semantics}, it is mainly caused by the weak discriminability of unseen class representations and remains a good topic for future ZSD research.
Notably, benefiting from the introduced cross-class representation regularization, our method achieves relatively good performances on many unseen classes(e.g. 20.0\% GZSD AP and 45.7\% ZSD AP for groundtrackfield class on DIOR). 

\subsection{Generalizability}
We further validate whether DescReg generalizes to the generative ZSD method and other detection architectures.

\paragraph{DescReg with Generative ZSD Methods
}
Generative methods aim at synthesizing samples for unseen classes, the proposed DescReg can be integrated into the framework for generating more discriminative samples. As shown in Tab.~\ref{tab:tab4_generative_pascal}, by augmenting with DescReg, the best-reporting generative method of RRFS is improved on PASCAL VOC dataset. Specifically, with DescReg, the mAP performance for unseen classes is improved from 65.5\% to 66.4\% on ZSD setting, and from 49.1\% to 50.4\% on GZSD setting.

\begin{table}[htb]
\centering
\begin{tabular}{c|c|ccc}
   \hline
   \multirow{2}*{\textbf{Method}} & \multirow{2}*{\textbf{ZSD}}
   & \multicolumn{3}{c}{\textbf{GZSD}} \\
   \cline{3-5}
   & & S & U & HM \\
   \hline  
   \rowcolor{lightgray!30} 
   SAN (2018) & 59.1 & 48.0 & 37.0 & 41.8\\
   HRE (2018) & 54.2 & \textbf{62.4} & 25.5 & 36.2\\
   \rowcolor{lightgray!30} 
   BLC (2020)& 55.2 & 58.2 & 22.9 & 32.9\\
   RRFS (2022)& 65.5 & 47.1 & 49.1 & 48.1\\
   \rowcolor{lightgray!30} 
   \makecell[c]{RRFS w/. DescReg} & \textbf{66.4} & 47.1 & \textbf{50.4} & \textbf{48.6}  \\
   \hline
\end{tabular}
\caption{ZSD and GZSD performance of the generative method on the PASCAL VOC dataset.}
\label{tab:tab4_generative_pascal}
\end{table}

\paragraph{DescReg with Other Detection Architectures}
In addition to Faster R-CNN, we further validate DescReg on the one-stage YOLOv8~\cite{Jocher_YOLO_by_Ultralytics_2023} and the multi-stage Cascaded R-CNN~\cite{cai2018cascade}. The results are shown in Tab.~\ref{tab:tab5_other_det_archs}. Our method applies well to the two detection models, e.g. with 15.6\% ZSD mAP and Cascased R-CNN and 6.4\% ZSD mAP on YOLOv8 which achieves 64 FPS inference speed.

\begin{table}[htb]
\centering
\begin{tabular}{c|c|ccc|c}
   \hline
   \multirow{2}*{\textbf{Architecture}} & \multirow{2}*{\textbf{ZSD}}
   & \multicolumn{3}{c}{\textbf{GZSD}}\vline & \multirow{2}*{\textbf{FPS}}\\
   \cline{3-5}
   & & S & U & HM & \\
   \hline  
   \rowcolor{lightgray!30} 
   Faster-RCNN & 15.2 & 68.7 & 7.9 & 14.2 & 11\\
   Cascaded-RCNN & \textbf{15.6} & \textbf{70.0} & \textbf{8.1} & \textbf{14.5} & 8\\
   \rowcolor{lightgray!30} 
   YOLOv8-s & 6.4 & 49.9 & 4.2 & 7.7 & \textbf{64}\\
   \hline
\end{tabular}
\caption{Performance with different detection architectures on the DIOR dataset.  FPS denotes frame per seconds.}
\label{tab:tab5_other_det_archs}
\end{table}

\subsection{Analysis}
We conduct several ablation studies and experimental analyses to better understand how the proposed method works. Please refer to supplementary for qualitative results.

\begin{figure}
\centering
\begin{tikzpicture}
\begin{axis}[
    width=0.39\textwidth,
    height=0.275\textwidth,
    extra description/.code={
        \draw [color=blue, line width=1pt] (axis cs:\pgfkeysvalueof{/pgfplots/xmin},\pgfkeysvalueof{/pgfplots/ymax}) -- (axis cs:\pgfkeysvalueof{/pgfplots/xmin},\pgfkeysvalueof{/pgfplots/ymin});
        \draw [color=red, line width=1pt] (axis cs:\pgfkeysvalueof{/pgfplots/xmax},\pgfkeysvalueof{/pgfplots/ymax}) -- (axis cs:\pgfkeysvalueof{/pgfplots/xmax},\pgfkeysvalueof{/pgfplots/ymin});
    },
    xlabel=Epoch,
    xlabel style={yshift=4.5pt},
    ylabel=mAP,
    ylabel style={yshift=-15pt}, 
    xmin=0,
    xmax=13,
    ymin=0,
    ymax=113,
    legend style={font=\fontsize{7}{7}\selectfont, at={(0.04,0.985)}, anchor=north west},
    ytick style={color=blue}
]
\addplot[
    color=blue,
    mark=square,
    ] coordinates {
    (0, 0)
    (1, 67.1)
    (2, 67.6)
    (3, 67.9)
    (4, 68.2)
    (5, 68.4)
    (6, 68.6)
    (7, 68.4)
    (8, 68.6)
    (9, 68.6)
    (10, 68.6)
    (11, 68.8)
    (12, 68.6)

};
\addlegendentry[text width=1.35cm]{seen w DR}
\addplot[
    color=blue,
    mark=triangle,
    ] coordinates {
    (0, 0)
    (1, 52.6)
    (2, 53.1)
    (3, 52.6)
    (4, 52.5)
    (5, 50.0)
    (6, 50.5)
    (7, 47)
    (8, 48.5)
    (9, 47)
    (10, 43.7)
    (11, 42.2)
    (12, 44.5)

};
\addlegendentry[text width=1.35cm]{seen w/o DR}

\end{axis}

\begin{axis}[
    width=0.39\textwidth,
    height=0.275\textwidth,
    extra description/.code={
        \draw [color=blue, line width=1pt] (axis cs:\pgfkeysvalueof{/pgfplots/xmin},\pgfkeysvalueof{/pgfplots/ymax}) -- (axis cs:\pgfkeysvalueof{/pgfplots/xmin},\pgfkeysvalueof{/pgfplots/ymin});
        \draw [color=red, line width=1pt] (axis cs:\pgfkeysvalueof{/pgfplots/xmax},\pgfkeysvalueof{/pgfplots/ymax}) -- (axis cs:\pgfkeysvalueof{/pgfplots/xmax},\pgfkeysvalueof{/pgfplots/ymin});
    },
    axis y line*=right,
    xmin=0,
    xmax=13,
    ymin=0,
    ymax=11.3,
    hide x axis,
    legend style={font=\fontsize{7}{7}\selectfont, at={(0.49,0.985)}, anchor=north west},
    ytick style={color=red}
]
\addplot[
    color=red,
    mark=square,
    ] coordinates {
    (0, 0.0)
    (1, 2.9)
    (2, 3.1)
    (3, 3.3)
    (4, 6.4)
    (5, 6.0)
    (6, 6.2)
    (7, 6.9)
    (8, 7.0)
    (9, 7.2)
    (10, 6.8)
    (11, 7.2)
    (12, 6.9)
};
\addlegendentry[text width=1.6cm]{unseen w DR}
\addplot[
    color=red,
    mark=triangle,
    ] coordinates {
    (0, 0)
    (1, 2.8)
    (2, 0.7)
    (3, 0.9)
    (4, 0.9)
    (5, 1.0)
    (6, 0.9)
    (7, 0.7)
    (8, 0.7)
    (9, 0.6)
    (10, 0.9)
    (11, 0.8)
    (12, 0.6)
};
\addlegendentry[text width=1.6cm]{unseen w/o DR}
\end{axis}
\end{tikzpicture}

\label{fig:multilineplot}
\caption{Learning dynamics of DescReg. w/wo DescReg denotes DescReg and baseline without DescReg.}
\end{figure}
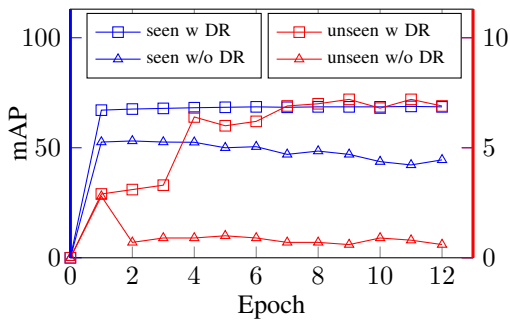

\paragraph{Ablation Study on the Proposed Triplet Loss}
As shown in Tab.~\ref{tab:tab6_ablation_triplet}, when replacing the semantic class embeddings with the visual description embeddings, the baseline performance is improved but the improvement is limited (e.g., 1.0\% on ZSD mAP). This result means naively incorporating the visual description information as the class semantic representation cannot help much due to the representation gap between semantic feature space and visual feature space. Additionally, by applying the proposed inter-class triplet loss, the performance is significantly improved (from 7.1\% to 10.1\% on ZSD mAP) which indicates that simple similarity-based triplet regularization could improve the zero-shot detection performance. By further introducing the proposed similarity-aware margin, the ZSD mAP is improved by 5.0\% and HM is improved by 5.6\%, meaning the adaptive margin helps better regularize the class representation space. We also observe the temperature value of 0.03 achieves the best performance, which is slightly higher than 0.01 and 0.05. Based on the best-performing model, further adding the visual description embeddings cannot offer improvement, indicating our method may already incorporate the visual characteristics into the embeddings through structural similarity regularization. Fig. 3 shows how the learning dynamics of seen and unseen classes, apparently, with DescReg, the  performance on both seen and unseen classes are higher and the learning process is more stable.

\begin{table}[htb]\centering
\fontsize{9.5pt}{10.5pt}\selectfont
\begin{tabular}{c|c|c|c|c}
\hline
\textbf{S$\rightarrow$V} & \textbf{Desc-Softmax} & \textbf{\makecell{Desc-Adaptive\\Margin}} & \textbf{ZSD} &  \textbf{HM} \\
\hline
\rowcolor{lightgray!30} 
 &  &  & 6.1 & 5.3 \\
 \checkmark &  &  & 7.1 & 5.9 \\
 \rowcolor{lightgray!30} 
 & \checkmark (0.03) &  & 10.1 & 8.2 \\
 & \checkmark (0.01) & \checkmark & 15.1 & 13.8 \\
 \rowcolor{lightgray!30} 
 & \checkmark (0.03) & \checkmark & 15.2 & \textbf{14.2} \\
 & \checkmark (0.05) & \checkmark & 14.5 & 13.6 \\
 \rowcolor{lightgray!30} 
\checkmark & \checkmark (0.03) & \checkmark & \textbf{15.3} & 13.9 \\
\hline
\end{tabular}
\caption{Ablation study of the proposed similarity-aware triplet loss. S$\rightarrow$V means replacing the semantic embeddings with visual description embeddings. Desc-Softmax and Desc-Adaptive-Margin denote the proposed self-excluding Softmax and the similarity-aware triplet loss. The numbers in the parentheses are the temperature used in the Softmax.}
\label{tab:tab6_ablation_triplet}
\end{table}

\paragraph{Effect of Varying Descriptions}
We investigate how sensitive is DescReg to the input visual descriptions by varying the description sources. We evaluate how different human and GPT-4~\cite{OpenAI2023} description inputs affect the zero-shot detection performance. As shown in Tab.~\ref{tab:tab7_desc_input_source}, when focusing on the semantics, the performance of both human and GPT-4 descriptions is low (e.g., 5.0\% ZSD mAP for human input and 6.9\% mAP for GPT-4 input). The reason is that simple semantic description contains much fewer visual details of the objects. When switching to descriptions that focus on visual details from an aerial view, the performance is significantly improved by more than 8.0\% in ZSD mAP and 7.0\% in HM, benefiting from the visual details that generate effective similarity measures. We also test how sensitive the method works with different descriptions of varying lengths. The result shows that our method is not very sensitive to the description length. In addition, we observe that GPT-generated descriptions offer higher performance than that of human inputs. While we did not dedicatedly optimize the human input, the result shows the application of large language models in ZSD is very efficient.

\begin{table}[htb]
\centering
\begin{tabular}{c|c|cc}
\hline
-                  & \textbf{Method} & \textbf{ZSD} & \textbf{HM} \\ \hline
 \rowcolor{lightgray!30} 
\multirow{2}{*}{\cellcolor{white}\textbf{Human}} & semantic & 5.0  & 4.7  \\ 
                           & aerial & 13.1  & 11.9  \\ \hline
                            \rowcolor{lightgray!30} 
                   \cellcolor{white} & semantic &  6.9  & 6.1  \\ 
                   & aerial-long  & \textbf{15.2}  & \textbf{14.2}  \\
                    \rowcolor{lightgray!30} 
                   \cellcolor{white}  & aerial-medium & 15.1  & 13.9  \\
                   \multirow{-4}{*}{\textbf{GPT-4}} & aerial-short & 13.5  & 13.2  \\
                   \hline
\end{tabular}

\caption{Varying descriptions. We acquire visual descriptions through both Human and GPT-4. \emph{semantic} means simply describing the object class while \emph{aerial} means focusing on the visual appearance in aerial images. \emph{long}, \emph{medium}, and \emph{short} denote descriptions with varying lengths.}
\label{tab:tab7_desc_input_source}
\end{table}

\subsection{Conclusion}
In this paper, we investigate the zero-shot object detection (ZSD) problem in the context of aerial images. We identified the weak semantic-visual correlation problem of aerial objects and propose to learn stronger visually-aligned class representations with external visual descriptions in text format. 
Our method is extensively validated on three challenging aerial object detection datasets and shows significantly improved performance to the prior ZSD methods. To the best of our knowledge, we are the first to conduct a comprehensive study on zero-shot aerial object detection.
we hope our method and newly established experimental setups provide a baseline for future research.
\paragraph{Limitations and Future Work} 
While our method significantly improves the baselines, we note the performance on unseen classes is still low. The major challenge arises from the strong inter-class confusion and background confusion among aerial objects, which is further exacerbated by the small object size. While our method mitigates these problems, two future directions could further address them: 1) The non-uniform spatial processing approaches~\cite{recasens2018learning,yang2019clustered} could be explored to amplify the small object signal for improved zero-shot recognition. 2) Based on our proposed regularization, other label-efficient methods could be incorporated to improve the performance, e.g. few-shot approach and open-vocabulary detection approach~\cite{kang2019few,wang2023learning}.

\section{Acknowledgments}
This work is supported by the Key Program of National Science Foundation of China under Grant 61836006, the Fundamental Research Funds for the Central Universities under Grant YJ202342 and 1082204112364, the National Science Foundation of China under Grant 62106161, the Key R\&D Program of Sichuan Province under Grant 2022YFN0017 and 2023YFG0278 and Engineering Research Center of Machine Learning and Industry Intelligence, Ministry of Education.

\bibliography{aaai24}

\clearpage

\begin{appendices}

\section{Appendix}

This appendix is organized as follows:
\begin{itemize}
    \item In Section A, we provide more dataset details about three remote sensing (aerial) datasets including image cropping, seen/unseen split and evaluation protocols details.
    \item In Section B, we provide more implementation details of our DescReg model.
    \item In Section C, we provide additional ablation studies, including the ablation studiews of more baselines for DescReg, different projection head architecture, different backbone depth and different embedding methods.
\end{itemize}

\section{A. More Dataset Details}

The details of the three remote sensing (aerial) datasets are as follows:
\begin{itemize}
    \item \textbf{DIOR}~\cite{Li2019ObjectDI} is a large-scale aerial object detection dataset with large-range object size variation and diverse visual qualities. It consists of 5862 train images, 5863 validation images, and 11725 test images. Total number of classes is 20. The images are in resolution 800x800. Following RRFS~\cite{Huang2022RobustRF}, we employ the combination of train and validation sets to the model and evaluate the model on the test set.
    \item \textbf{xView}~\cite{Lam2018xViewOI} is a high-resolution aerial object detection dataset, it is collected from WorldView-3 satellites at 0.3m ground sample distance. The resolution is in the range of about 2500×2500 to about 5000×3500 pixels.
    The dataset consists of 846 annotated images with 60 classes, of which we sample 665 images for training and 181 images for evaluation. 
    \item \textbf{DOTA} ~\cite{Xia2017DOTAAL} is collected from multiple sensors and platforms such as Google Earth. The resolution is about 800×800 to about 4,000×4,000 pixels. We employ the 1.0 version, which consists of 1411 training images, and 458 validation images. There are 15 categories annotated in the dataset. Models are trained on the train set and evaluated on the validation set.

\end{itemize}

\subsection{Image Cropping}
For DIOR, we directly employ the images for ZSD experiments following prior work~\cite{Huang2022RobustRF}. For xView and DOTA, due to their varying image size, we simplify the dataset by conducting cropping preprocessing. Each image in the datasets is further cropped into images of size 800 * 800. 
Concretely, we crop the image horizontally to $ w//800 + 1$ patches. The number of overlapping pixels is $800+\frac{800-w}{w//800}$, the patches are thus evenly distributed along the width dimension. In practice, we find some of the DOTA images are less than 800 in size, thus if $w \leq 800$, the image is not cropped and we pad the width to 800. 
The same cropping procedure is done for the image height dimension. 
With the cropping, we obtain 12826/3549 and 18430/6259 images (training/test) for Xview and DOTA, respectively.

\subsection{Seen/unseen Split} For DIOR, we follow RRFS~\cite{Huang2022RobustRF} to split the 20 classes in to 16/4 for seen/unseen classes.
For Xview and DOTA, to maintain unseen class diversity (i.e., the unseen classes should be diverse in semantics) and semantic relatedness (i.e., each unseen class should have at least one semantically close seen class.), \emph{we perform hierarchical clustering on the class semantic embeddings and sample one class for pairs of leaf nodes in the clustering tree.} For example, 'ground-track-field' and 'soccer-ball-field' are clustered together and the soccer-ball-field is sampled as unseen classes. The resulting splits are 11/4 and 48/12 for DOTA and xView respectively. These class splits are:
\begin{itemize}
    \item DIOR: \textbf{seen}: 'airplane', 'baseballfield', 'bridge', 'chimney', 'dam', 'Expressway-Service-area', 'Expressway-toll-station', 'golffield', 'harbor', 'overpass', 'ship', 'stadium', 'storagetank', 'tenniscourt', 'trainstation', 'vehicle', 
    \textbf{unseen}: 'airport', 'basketballcourt', 'groundtrackfield', 'windmill'.
    \item xView: \textbf{seen}: 'fixed wing aircraft', 'small aircraft', 'passenger plane or cargo plane', 'passenger vehicle', 'small car', 'utility truck', 'truck', 'cargo truck', 'truck tractor', 'trailer', 'truck tractor with flatbed trailer', 'truck tractor with liquid tank', 'crane truck', 'railway vehicle', 'passenger car', 'cargo car or container car', 'flat car', 'tank car', 'locomotive', 'sailboat', 'tugboat', 'fishing vessel', 'ferry', 'yacht', 'container ship', 'oil tanker', 'engineering vehicle', 'tower crane', 'container crane', 'straddle carrier', 'dump truck', 'haul truck', 'front loader or bulldozer', 'cement mixer', 'ground grader', 'hut or tent', 'shed', 'building', 'aircraft hangar', 'damaged building', 'facility', 'construction site', 'vehicle lot', 'helipad', 'storage tank', 'shipping container', 'pylon', 'tower'. \textbf{unseen}: 'helicopter', 'bus', 'pickup truck', 'truck tractor with box trailer', 'maritime vessel', 'motorboat', 'barge', 'reach stacker', 'mobile crane', 'scraper or tractor', 'excavator', 'shipping container lot'.
    \item DOTA: \textbf{seen}: 'plane', 'ship', 'storage-tank', 'baseball-diamond', 'basketball-court', 'ground-track-field', 'harbor', 'bridge', 'large-vehicle', 'small-vehicle', 'roundabout'. \textbf{unseen}: 'tennis-court', 'helicopter', 'soccer-ball-field', 'swimming-pool'.
\end{itemize}

\begin{figure*}[htbp]   
	\centering
	\includegraphics[width=\linewidth,scale=1.00]{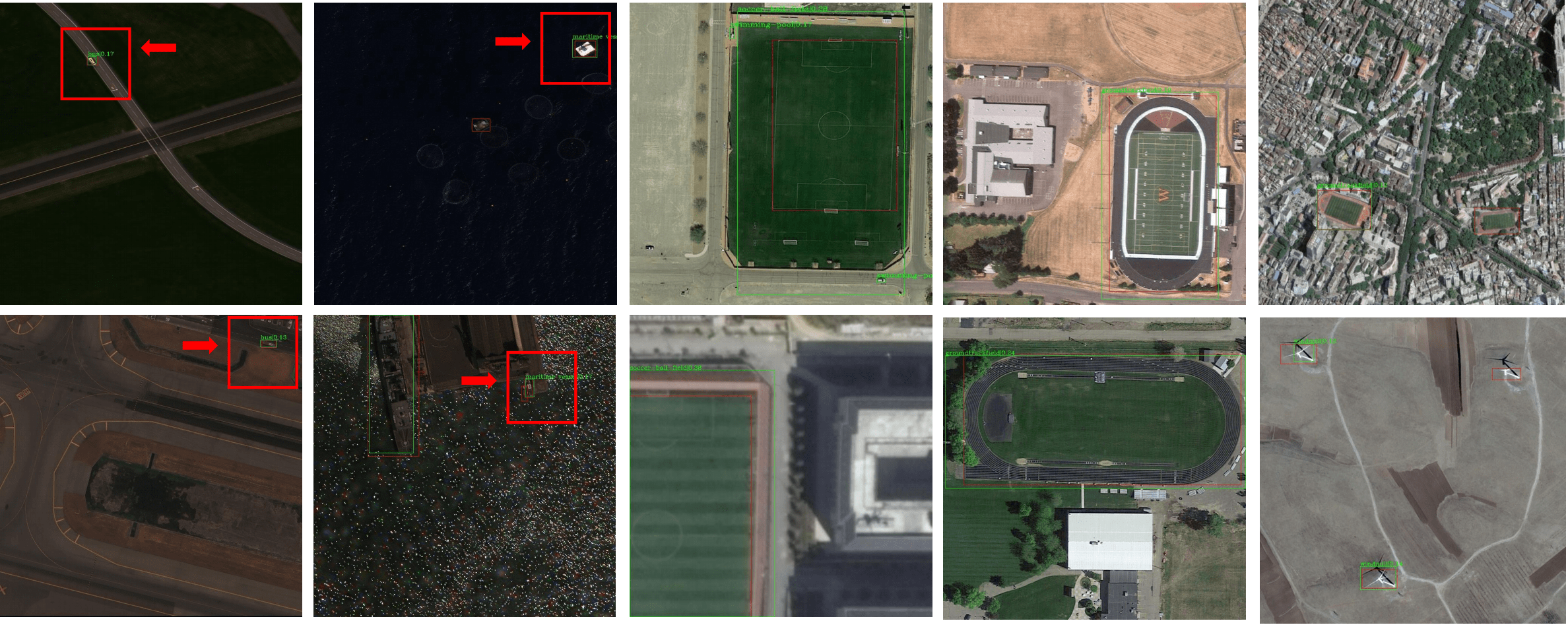}
	\caption{Qualitative results on DIOR, xView and DOTA, for unseen objects. The green are detections and the red are ground-truth boxes. The small results on xView are marked with red anchors. Best viewed with zoom-in.}
	\label{FigureOne}
\end{figure*}

\subsection{Evaluation Protocols} Same as common ZSD setup~\cite{Bansal2018ZeroShotOD}, we employ Recall@100 (RE@100) and mean average precision (mAP) as the criteria for evaluation with an intersection over union (IoU) of 0.5. 
In the GZSD setting, the performance is evaluated on both seen and unseen subsets. The Harmonic Mean (HM)~\cite{Han2021ContrastiveEF} is used as a final metric to combine seen and unseen perfomrance.
For the ZSD setting, we follow prior works~\cite{Huang2022RobustRF,yan2022semantics,rahman2020improved} and evaluate on test images that only contain unseen classes. For the first 6 unseen classes on Xview (i.e., 'helicopter', 'bus', 'pickup truck', 'truck tractor with box trailer', 'maritime vessel', 'motorboat'), this criterion filters too many images and keeps only a few of ground truth, which results in exceptionally high AP results. We thus relax this image selection criterion and include images that contain the least seen class objects for the first 6 unseen classes.

\section{B. More Implementation Details}

\subsection{DescReg Implementation} 
\paragraph{Training} As mentioned in our main paper, we conduct two-stage training: 1) In the first stage, the model is trained on the seen class data, and the training takes 12 epochs on all three datasets, the learning rate is 0.01 with decay at 8th and 11th epochs. We employ SGD with $momentum=0.9$ and $weight-decay=0.0001$ as the optimizer.
2) In the second stage, the model is frozen and the classifier layer is replaced with the projection head, the projection head is fine-tuned with DescReg objective, which consists of classification loss and triplet loss. The optimizer detail is the same as the first stage of training, with only a different learning rate of 0.02. For background embedding, We follow the prior work of BLC~\cite{Zheng2020BackgroundLC} and employ learnable background representation during the second-stage training.

\paragraph{Implementation on RRFS}
To incorporate DescReg into RRFS~\cite{Huang2022RobustRF} and compare with it on the Pascal VOC dataset, we first generate descriptions on the 20 classes with GPT-4 and then follow the same process as on the aerial dataset to compute the triplet loss, the original inter-class contrastive loss in RRFS is then replaced with our triplet objective. The other training sets and hyper-parameters are identical to RRFS.

\paragraph{Implementation on Cascaded R-CNN and YOLOv8}
It is very simple to integrate our method into other object detection meta-architectures such as cascaded object detectors and one-stage detection detectors.
We employ Cascaded R-CNN~\cite{cai2018cascade} and YOLOv8~\cite{Jocher_YOLO_by_Ultralytics_2023} to validate the generalization on these architectures.
For Cascaded R-CNN, after the first stage detection training on seen class data, we replace the classifier of each stage with an independent semantic projection head and conduct fine-tuning with the proposed DescReg. For YOLOv8, unlike Faster R-CNN and Cascaded R-CNN which employ Softmax for classification by default, it uses class-wise Sigmoid loss for classification, we simply replace the logit before Sigmoid activation with dot product similarity between the projected class embeddings and visual features. The training follows the same two-stage fashion as Faster R-CNN and Cascaded R-CNN.


\section{C. Additional Ablation Studies}

\subsection{More Baselines for DescReg}
We evaluate some other alternative Design Choices of DescReg as baselines: 1)forcing diagonal similarity matrix in triplet loss regularization, i.e., each class is only similar to itself and is not similar to any other classes (0.0 similarity). In this way, the proposed triplet loss is reduced to an inter-class contrastive loss.
2)direct similarity regularization: we compute cosine embedding similarity and 
directly impose L2 loss with the similarity computed from the descriptions.
As shown in Tab.~\ref{tab:tab10_alternative_designs}, these two baseline alternative choices generate much worse results compared to the proposed similarity-aware triplet loss formulation. This results shows that our method can more effectively encode the structural inter-class similarity information into the embedding learning and thus achieves better ZSD performance.

\begin{table}[htb]
\centering
\begin{tabular}{c|c|c}
\hline
- & \textbf{ZSD} & \textbf{HM} \\
\hline
\rowcolor{lightgray!30} 
\textbf{proposed} & \textbf{15.2} & \textbf{14.2} \\
\hline
\textbf{diagonal similarity} & 7.3 & 6.4 \\
\hline
\rowcolor{lightgray!30} 
\textbf{direct similarity reg} & 8.7 & 6.2 \\
\hline
\end{tabular}
\caption{Alternative Design Choices for DescReg. \emph{diagonal similarity} means forcing the similarity matrix as a diagonal eye matrix. \emph{direct similarity reg} denotes imposing L2 loss directly on the cosine similarity computed from the embedding and descriptions.}
\label{tab:tab10_alternative_designs}
\end{table}

\subsection{Head Architecture} Tab.~\ref{tab:tab9_projection_head} shows an ablation study on the projection head architecture. We thus employ a 1-layer projection head for DIOR and xView, and a 2-layer projection head for DOTA. The results indicate that DesReg performance depends on the projection head architecture as well as the dataset. It is better to validate and find the best setting.

\begin{table}[htb]
\centering
\setlength{\abovecaptionskip}{5pt}
\renewcommand{\arraystretch}{1.2}
\begin{tabular}{c|c|cc}
\hline
-                  & \textbf{Arch} & \textbf{ZSD} & \textbf{HM} \\ \hline
\rowcolor{lightgray!30} 
 \cellcolor{white} & 1-fc & \textbf{15.2} & \textbf{14.2} \\ 
                           & 2-fc & 14.8 & 13.7 \\ 
                           \rowcolor{lightgray!30}  
                           \cellcolor{white}\multirow{-3}*{\textbf{DIOR}} & 3-fc &  12.5 & 11.0  \\ 
                           \hline
 \cellcolor{white} & 1-fc & 8.0 & 8.1 \\ 
                           & 2-fc & \textbf{8.5} & \textbf{8.8} \\ 
                           \rowcolor{lightgray!30}  
                           \cellcolor{white}\multirow{-3}*{\textbf{DOTA}} & 3-fc &  7.2 & 7.3  \\ 
                           \hline
                   \rowcolor{lightgray!30}
 \cellcolor{white} & 1-fc &  \textbf{8.3}  & \textbf{8.7}  \\  
                   \cellcolor{white} & 2-fc & 7.7  & 7.5    \\
                   \rowcolor{lightgray!30}
                   \cellcolor{white}\multirow{-3}{*}{\textbf{Xview}} & 3-fc & 6.7  & 6.3  \\
                   \hline
\end{tabular}
\caption{Effect of different projection head architecture, if the layer is more than 1, we use intermediate feature dimension of 128. The experiment is conducted on DIOR dataset.}
\label{tab:tab9_projection_head}
\end{table}

\subsection{Different Backbone Depth}
As shown in Tab.~\ref{tab:tab11_backbone}, we replace the default ResNet-101 backbone with ResNet-50 and ResNet-152. The results indicate that the performance improves significantly from ResNet-50 to ResNet-101 (e.g., 12.3 to 15.2 in ZSD mAP). However, further increasing the backbone depth to ResNet-152 obtains diminished gain on unseen class detection. 
While the seen class mAP is improved with the deeper network, the result shows that the unseen class may not benefit further with deeper network representation.

\begin{table}[htb]
\centering
\begin{tabular}{c|c|c|c}
\hline
- & \textbf{seen} & \textbf{ZSD} & \textbf{HM} \\
\hline
\rowcolor{lightgray!30} 
\textbf{res-50} & 60.5 & 12.3 & 10.9 \\
\hline
\textbf{res-101} & 68.7 & \textbf{15.2} & 14.2 \\
\hline
\rowcolor{lightgray!30} 
\textbf{res-152} & \textbf{70.1} & 15.0 & \textbf{14.4} \\
\hline
\end{tabular}
\caption{The effect of backbone depth. The experiments are conducted on DIOR dataset. \emph{seen} denotes mAP on seen classes.}
\label{tab:tab11_backbone}
\end{table}

\paragraph{Embedding Models}
We examine the effect of different semantic embedding methods and visual embedding methods. As shown in Tab.~\ref{tab:tab8_embedding_models}, we find: 1) For both semantic embeddings and description embeddings, the pre-trained language models such as BERT offer relatively good results, e.g., 15.2\% ZSD mAP with BERT semantic embedding and S-BERT description embedding. 2) The vision-language pre-trained model of CLIP~\cite{radford2021learning} can further improve the best-performing language model (e.g., 15.8\% vs. 15.2\% with CLIP semantic embedding and BERT semantic embedding), this is likely due to the leaked visual cues in the vision-language representation. We mainly adopt pure language models to conform with the zero-shot setting.

\begin{table}[htb]
\centering
\begin{tabular}{c|c|cc}
\hline
-                  & \textbf{Method} & \textbf{ZSD} & \textbf{HM} \\ \hline
\rowcolor{lightgray!30} 
 \cellcolor{white} & Word2Vec (2013) & 12.0  & 11.7  \\ 
                           & GloVe (2014) & 12.1  & 10.9  \\ 
                           \rowcolor{lightgray!30} 
                           \cellcolor{white} & BERT (2019) & 15.2  & 14.2  \\ 
                           \multirow{-4}{*}{\textbf{Semantic}} & CLIP (2021) &  \textbf{15.8} & \textbf{14.6}  \\ 
                           \hline
                           \rowcolor{lightgray!30} 
 \cellcolor{white} & BERT (2018) &  13.5  & 13.1  \\ 
                   & DSG (2018)  & 14.9  & 13.6  \\
                   \rowcolor{lightgray!30} 
                   \cellcolor{white} & S-BERT (2019) & 15.2  & 14.2    \\
                   \multirow{-4}{*}{\textbf{Description}} & CLIP (2021) & \textbf{15.5}  & \textbf{14.6}  \\
                   \hline
\end{tabular}
\caption{Effect of different embedding methods. The results are obtained for semantic class names and visual descriptions independently, i.e., by fixing one and varying the other.}
\label{tab:tab8_embedding_models}
\end{table}

\paragraph{Qualitative Results}
As shown in Fig.~\ref{FigureOne} is some example qualitative results, our method can correctly detect some challenging unseen objects such as the windmill, bus, and maritime vessel.

\end{appendices}

\end{document}